\documentclass{article}

\PassOptionsToPackage{numbers, compress}{natbib}
\usepackage[final]{ARLET_2024}

\usepackage[utf8]{inputenc} % allow utf-8 input
\usepackage[T1]{fontenc}    % use 8-bit T1 fonts
\usepackage[hidelinks]{hyperref}       % hyperlinks
\usepackage{url}            % simple URL typesetting
\usepackage{booktabs}       % professional-quality tables
\usepackage{amsfonts}       % blackboard math symbols
\usepackage{nicefrac}       % compact symbols for 1/2, etc.
\usepackage{microtype}      % microtypography
\usepackage[dvipsnames]{xcolor}

\usepackage{booktabs}
\usepackage{amsmath}
\usepackage{wrapfig}

\newcommand{\cvec}[1]{\boldsymbol{\mathrm{#1}}}
\newcommand{\cmat}[1]{\boldsymbol{\mathrm{#1}}}
\newcommand{\KL}[2]{\textrm{KL}\left[ {#1} \parallel {#2} \right]}

\newcommand{\ours}{\emph{KalMamba}}

\title{KalMamba: Towards Efficient Probabilistic State Space Models for RL under Uncertainty}

\author{%
Philipp Becker\thanks{Correspondence to \url{philipp.becker@kit.edu}.} 
\And
Niklas Freymuth\\
Karlsruhe Institute of Technology 
\And
Gerhard Neumann
}

\usepackage{tikz}
\usetikzlibrary{arrows.meta, shapes.geometric, positioning, backgrounds}

\newcommand\blfootnote[1]{%
  \begingroup
  \renewcommand\thefootnote{}\footnote{#1}%
  \addtocounter{footnote}{-1}%
  \endgroup
}

\begin{document}

\maketitle

\begin{abstract}
\blfootnote{
Versions of this work were also presented at the \textit{Next Generation of Sequence Modeling Architectures} workshop at ICML 2024 and the \textit{Training Agents with Foundation Models} workshop at RLC 2024}
%Latent representations from probabilistic State Space Models (SSMs) are essential for Reinforcement Learning (RL) from high-dimensional, partial information.
Probabilistic State Space Models (SSMs) are essential for Reinforcement Learning (RL) from high-dimensional, partial information as they provide concise representations for control. 
Yet, they lack the computational efficiency of their recent deterministic counterparts such as~\emph{S4} or \emph{Mamba}.
We propose \emph{KalMamba}, an efficient architecture to learn representations for RL that combines the strengths of probabilistic SSMs with the scalability of deterministic SSMs. 
\emph{KalMamba} leverages \emph{Mamba} to learn the dynamics parameters of a linear Gaussian SSM in a latent space. 
Inference in this latent space amounts to standard Kalman filtering and smoothing.
We realize these operations using parallel associative scanning, similar to \emph{Mamba}, to obtain a principled, highly efficient, and scalable probabilistic SSM.
Our experiments show that \emph{KalMamba} competes with state-of-the-art SSM approaches in RL while significantly improving computational efficiency, especially on longer interaction sequences.
%Probabilistic state space models (SSMs) play a crucial role in RL from high-dimensional, partial information, but lack the computational efficiency of their recent deterministic counterparts for long sequences. 
%We propose \emph{KalMamba}, an efficient architecture for Reinforcement Learning (RL) that merges the capabilities of probabilistic SSMs with the beneficial scaling properties of deterministic SSMs. 
%\emph{KalMamba} builds on \emph{Mamba} to learn the dynamics parameters of a linear Gaussian SSM in a latent space.
%Inference in this latent space can be done using standard Kalman filtering and smoothing, which is amenable to parallel scanning, similar as \emph{Mamba} itself. 
%This approach ensures computational efficiency while appropriately handling uncertainties by considering both past and future observations. 
%Preliminary experiments demonstrate that \emph{KalMamba} performs competitively with other state-of-the-art state-space approaches for RL, while significantly enhancing computational efficiency.
\end{abstract}

 \section{Introduction}

Deep probabilistic State Space Models (SSMs) are versatile tools widely used in Reinforcement Learning (RL) for environments with high-dimensional, partial, or noisy observations~\cite{hafner2019dream, lee2020stochastic, nguyen2021tpc,becker2022uncertainty,hafner2023mastering,samsami2024mastering}.
They model states and observations as random variables and relate them through a set of conditional distributions, allowing them to capture uncertainties and learn concise probabilistic representations for downstream RL applications. 
Beyond RL, recent deterministic SSMs~\cite{gu2021s4, smith2022simplified, gu2023mamba} offer a powerful new paradigm for general sequence modeling and rival state-of-the-art transformers while improving computational complexity~\cite{gu2023mamba}.
These models assume states and observations are vectors related by deterministic, linear, and associative functions, which allow efficient time-parallel computations.
Such deterministic models are often insufficient for RL with complex observations, where uncertainty awareness and probabilistic modeling are crucial~\cite{chua2018pets, lee2020stochastic, hafner2019learning}.
In contrast, due to their nonlinear parameterizations and inference approaches, most existing probabilistic SSMs for RL and beyond do not feature the favorable scaling behavior of recent deterministic SSMs.

Many real-world applications require both uncertainty awareness and the capability of handling long sequences.
Examples include multi-modal robotics tasks with high-frequency control, long sequence non-stationary tasks, or complex information-gathering tasks. 
Consider a robot tasked with packing objects of unknown properties into a basket. 
By interacting with each item to infer and memorize properties such as mass and deformability, the robot refines its understanding of the scene, enabling it to strategically arrange the objects in the basket.
Current deterministic SSMs lack uncertainty awareness to solve such tasks, while their probabilistic counterparts do not scale to the required sequence lengths.
Thus, the question of how to develop a principled method that combines the benefits of both paradigms to obtain robust and efficient probabilistic state space models for long-sequence RL under uncertainty arises. 

\begin{figure}[t]
    \centering
     \includegraphics[width=\textwidth]{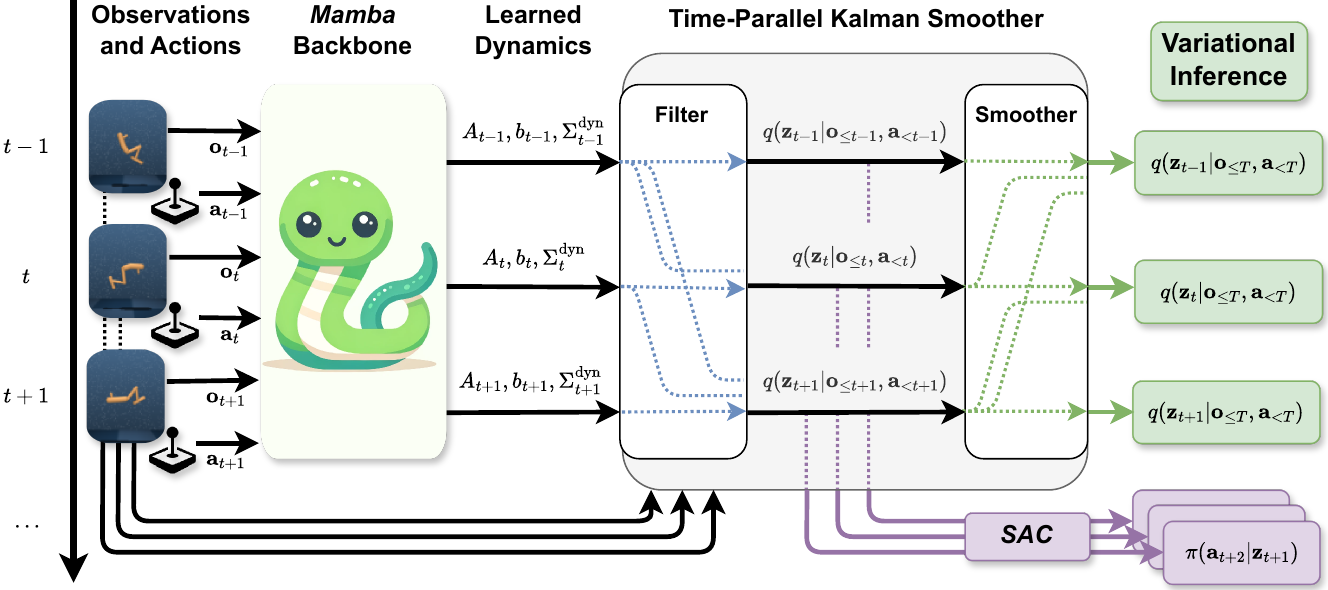}
    \caption{
    Overview of \ours. The observation-action sequences are first fed through a dynamics backbone built on \emph{Mamba}~\citep{gu2023mamba} to learn a linear dynamics model for each step.
    \ours~then uses time-parallel Kalman filtering~\cite{sarkka2020temporal} to infer filtered beliefs $q(\cvec{z}_t | \cvec{o}_{\leq t}, \cvec{a}_{\leq t-1})$ which can be used for control with a \emph{Soft Actor Critic (SAC)}~\cite{haarnoja2018sac}. 
    For model training, \ours~employs an additional time-parallel Kalman smoothing step to obtain smoothed beliefs $q(\cvec{z}_t | \cvec{o}_{\leq T}, \cvec{a}_{\leq T})$.
    These beliefs allow training a model that excels in modeling uncertainties due to a tight variational lower bound~\cite{becker2022uncertainty}. 
    Crucially, the smoothing step does not introduce trainable model parameters, enabling the direct use of the filtered beliefs for downstream RL policy training and execution.
    }
    \label{fig:fig1}
\end{figure}

We propose an efficient architecture for RL that equips probabilistic SSMs with the efficiency of recent deterministic SSMs.
Our approach, \emph{KalMamba}, uses (extended) Kalman filtering and smoothing~\cite{kalman1960filter, rauch1965maximum, jazwinski1970stochastic} to infer belief states over a linear Gaussian SSM in a latent space that uses a dynamics model based on \emph{Mamba}~\citep{gu2023mamba}.
In this approach, Mamba acts as a highly effective general-purpose sequence-to-sequence model to learn the parameters of a dynamics model. 
The Kalman Smoother uses this model to compute probabilistic beliefs over system states.
Figure~\ref{fig:fig1} provides a schematic overview.
\emph{Mamba} is efficient for long sequences as it uses parallel associative scans, which allow parallelizing associative operators on highly parallel hardware accelerators such as GPUs~\cite{sengupta2007scan}.
Similarly, we formulate both Kalman filtering and smoothing as associative operations~\cite{sarkka2020temporal} and build efficient parallel scans for filtering and smoothing in PyTorch~\cite{paszke2019pytorch}.  
With both \emph{Mamba} and the Kalman Smoother being parallelizable, \ours~achieves time-parallel computation of belief states required for model learning and control. 
Thus, unlike previous approaches for efficient SSM-based RL~\citep{samsami2024mastering}, which rely on simplified inference assumptions,~\emph{KalMamba} enables end-to-end model training under high levels of uncertainty using a smoothing inference and tight variational lower bound~\citep{becker2022uncertainty}.
While using smoothed beliefs for model learning, our architecture ensures a tight coupling between filtered and smoothed belief states.
This inductive bias ensures the filtered beliefs are meaningful, allowing their use for policy learning and execution where future observations are unavailable.

We evaluate~\ours~on several tasks from the DeepMind Control (DMC) Suite~\citep{tassa2018deepmind}, training an off-the-shelf \emph{Soft Actor-Critic}~\citep{haarnoja2018sac} on beliefs inferred from both images and states.
As baselines, we compare to \emph{Recurrent State Space Models}~\cite{hafner2019learning} and the \emph{Variational Recurrent Kalman Network}~\citep{becker2022uncertainty}.
Our preliminary experiments show that~\ours~is competitive to these state-of-the-art SSMs while being significantly faster to train and scaling gracefully to long sequences due to its ability to be efficiently parallelized.
These results indicate~\ours's potential for applications that require forming accurate belief states over long sequences under uncertainty.

To summarize our contributions, we
\mbox{(i) propose}~\ours, a novel probabilistic SSM for RL that combines Kalman filtering, smoothing, and a Mamba backbone to offer efficient probabilistic inference,
\mbox{(ii) motivate} and compare~\ours~to existing probabilistic SSMs for RL, and
\mbox{(iii) validate} our approach on state- and image-based control tasks, closely matching the performance of state-of-the-art probabilistic SSMs while being time-parallelizable.

% \newpage
 \section{Related Work}

\textbf{Deterministic State Space Models in Deep Learning.}
Structured deterministic State Space approaches~\cite{gu2021s4,smith2022simplified,gu2023mamba} recently emerged as an alternative to the predominant Transformer~\cite{vaswani2017attention} architecture for general sequence modeling~\cite{gu2023mamba}. 
Their main benefit is combining compute and memory requirements that scale linearly in sequence length with efficient and parallelizable implementations. 
While earlier approaches, such as the \emph{Structured State Space Sequence Model (S4)}~\cite{gu2021s4} and others~\cite{gupta2022dss,hasani2022liquid} used a convolutional formulation for efficiency, more recent approaches~\cite{smith2022simplified, gu2023mamba} use associative scans. 
Such associative scans allow for parallel computations over sequences if all involved operators are associative, which yields a logarithmic runtime, given enough parallel cores.
However, all these models are deterministic, i.e., they do not model uncertainties or allow sampling without further modifications.
As a remedy, \emph{Latent S4 (LS4)}~\cite{zhou2023ls4} extends \emph{S4} for probabilistic generative sequence modeling and forecasting. 
However, in LS4, the latent states are not Markovian and are thus hard to use for control. 
\emph{KalMamba} exploits the fact that filtering and smoothing in linear Gaussian state space models can also be formulated as a set of associative operations, which makes it amenable to parallel scans~\cite{sarkka2020temporal}. 
To our knowledge, it is the first deep-learning model to do so.
Further, it relies on \emph{Mamba}~\cite{gu2023mamba}, a state-of-the-art deterministic state space model, to precompute the dynamics models required for filtering and smoothing. 

\textbf{Probabilistic State Space Models for Reinforcement Learning.}
Probabilistic state space models are commonly and successfully used for reinforcement learning from high dimensional or multimodal observations~\cite{nguyen2021tpc, wu2022daydreamer, hafner2023mastering, becker2023joint}, under partial observability~\cite{becker2022uncertainty}, and for memory tasks~\cite{samsami2024mastering}. 
Arguably, the most prominent approach is the \emph{Recurrent State Space Model (RSSM)}~\cite{hafner2019learning}. 
After their original introduction as the basis of a standard planner, they have been improved with more involved parametric policy learning approaches~\cite {hafner2019dream} and categorical latent variables for categorical domains~\cite{hafner2020mastering}. 
During inference, the \emph{RSSMs} conditions the latent state on past observations and actions, resulting in a filtering inference scheme.
Here, the key architectural feature of \emph{RSSMs} is splitting the latent state into stochastic and deterministic parts. 
The deterministic part is then propagated through time using a standard recurrent architecture.
In its original formulation, the \emph{RSSM} uses a Gated Recurrent Unit (GRU)~\cite{cho2014gru}. 
One line of research focuses on replacing this deterministic path with more efficient architectures with the \emph{TransDreamer}~\cite{chen2022transdreamer} approach using a transformer~\cite{vaswani2017attention} and \emph{Recall to Image}~\cite{samsami2024mastering} using S4~\cite{gu2021s4}. 
However, to fully exploit the efficiency of these backbone architectures, both need to simplify the inference assumptions and can only consider the current observation, which makes them highly susceptible to noise or missing observations. 
Opposed to that, the \emph{Variational Recurrent Kalman Network (VRKN)}~\cite{becker2022uncertainty} proposes using a smoothing inference scheme that conditions both past and future actions. 
This scheme allows the \emph{VRKN} to work with a fully stochastic latent state and lets it excel in tasks where modeling uncertainty is crucial. 
The \emph{VRKN} uses a locally linear Gaussian State Space Model in a latent space, performing closed-form Kalman Filtering and smoothing. 
\emph{KalMamba} holistically combines smoothing inference in a
fully probabilistic SSM with an efficient temporally parallelized implementation, resulting in an approach that is robust to noise and efficient.

\textbf{Probabilistic State Space Models in Deep Learning.} 
%{\color{red}
Probabilistic state space models are versatile and commonly used tools in machine learning. 
Besides classical approaches using linear models \cite{shumway1982approach} and works using Gaussian Processes \cite{eleftheriadis2017gpssm, doerr2018prssm}, most recent methods build on Neural Networks (NNs) to parameterize generative and inference models using the SSM assumptions~\cite{archer2015black, watter2015embed, gu2015NASMC, karl2016dvbf, fraccaro2017kvae,  
krishnan2017sin, banijamali2018robust, yingzhen2018dsa, schmidt2018dssm_flow, naesseth2018vsmc, becker2019recurrent,becker-ehmck2019sld_dvbf,moretti2019svsmc,shaj2020action,klushyn2021latent,shaj2022hidden}.
Out of these approaches, those that embed linear-Gaussian SSMs into latent spaces~\cite{watter2015embed,  haarnoja2016backprop, fraccaro2017kvae, banijamali2018robust, becker-ehmck2019sld_dvbf, becker2019recurrent, shaj2020action, klushyn2021latent, shaj2022hidden} are of particular relevance to \emph{KalMamba}. 
Doing so allows for closed-form inference using (extended) Kalman Filtering and Smoothing. 
However, with the notable exception of the \emph{VRKN}, these models usually cannot be used to control or even model systems of similar complexity to those controlled with \emph{RSSM}-based approaches. 
Furthermore, some of them \cite{karl2016dvbf, becker-ehmck2019sld_dvbf} do not allow smoothing, while others \cite{fraccaro2017kvae, klushyn2021latent} model observations in the latent space as additional random variables which complicates inference and training and prevents principled usage of the observation uncertainty for filtering. 
Another class of approaches~\cite{haarnoja2016backprop, becker2019recurrent, shaj2020action,shaj2022hidden} trains using regression and are thus not generative. 
Notably, none of these approaches uses a temporally parallelized formulation of the filtering and smoothing operations.%}
\emph{KalMamba} takes inspiration from many of these approaches and partly follows the \emph{VRKN}'s design to enable reinforcement learning for complex systems. 
However, it combines those ideas with the efficiency of recent deterministic SSMs using an architecture that enables time-parallel computations.

% \newpage
 \section{State Space Models for Reinforcement Learning}
In Reinforcement Learning (RL) under uncertainty and partial observability, State Space Models (SSMs) generally assume sequences of observations $\cvec{o}_{\leq T} = \lbrace \cvec{o}_t \rbrace_{t = 0 \cdots T}$ which are generated by a sequence of latent state variables $\cvec{z}_{\leq T} = \lbrace \cvec{z}_t\rbrace_{t = 0 \cdots T}$, given a sequence of actions $\cvec{a}_{\leq T} = \lbrace \cvec{a}_t\rbrace_{t = 0 \cdots T}$. 
The corresponding generative model factorizes according to the hidden Markov assumptions~\cite{murphy2012machine},
i.e., each observation $\cvec{o}_t$  only depends on the current latent state $\cvec{z}_t$ through an observation model $p(\cvec{o}_t | \cvec{z}_t)$, and each latent state $\cvec{z}_t$ only depends on the previous state $\cvec{z}_{t-1}$ and the action $\cvec{a}_{t-1}$ through a dynamics model $p(\cvec{z}_t | \cvec{z}_{t-1}, \cvec{a}_{t-1})$. 

In order to learn the state space model from data and use it for downstream RL, we need to infer latent belief states given observations and actions.
Depending on the information provided for inference, we differentiate between the filtered belief $\cvec q(\cvec{z}_t | \cvec{o}_{\leq t}, \cvec{a}_{\leq t-1})$ and the smoothed belief $\cvec q(\cvec{z}_t | \cvec{o}_{\leq T}, \cvec{a}_{\leq T})$.
The filtered belief conditions only on past information, while the smoothed belief also depends on future information. 
Computing these beliefs is intractable for models of reasonable complexity. 
Thus, we resort to an autoencoding variational Bayes approach that allows joint training of the generative and an approximate inference model using a lower bound objective~\cite{kingma2013auto}.

\begin{wraptable}{R}{0.65\textwidth}
\vspace{-0.5cm}
\caption{Comparing the inference models and capabilities for smoothing (Smooth) and time-parallel (Parallel) execution of recent SSMs for RL. 
}
\vspace{0.2cm}
\centering
\begin{tabular}{lccc}
\toprule
Method & Inference Model & Smooth & Parallel \\ 
\midrule
RSSM~\cite{hafner2019learning} & $q(\cvec{z}_t | \cvec{h}_t, \cvec{o}_{t})$  &  {\color{BrickRed} $\boldsymbol{\times}$} & {\color{BrickRed} $\boldsymbol{\times}$}   \\
R2I~\cite{samsami2024mastering} & $q(\cvec{z}_t | \cvec{o}_t)$ & {\color{BrickRed} $\boldsymbol{\times}$} & {\color{ForestGreen} \checkmark}\\ 
VRKN~\cite{becker2022uncertainty} & $q(\cvec{z}_t | \cvec{o}_{\leq T}, \cvec{a}_{\leq T})$ & {\color{ForestGreen} \checkmark} & {\color{BrickRed} $\boldsymbol{\times}$}  \\
KalMamba &$q(\cvec{z}_t | \cvec{o}_{\leq T}, \cvec{a}_{\leq T})$& {\color{ForestGreen} \checkmark} & {\color{ForestGreen} \checkmark} \\ 
\bottomrule
\end{tabular}
\end{wraptable}

The \emph{Recurrent State Space Model (RSSM)}~\cite{hafner2019learning} assumes a nonlinear dynamics model, splitting the state $\cvec{z}_t$ into a stochastic $\cvec{s}_{t}$ and a deterministic part $\cvec{h}_{t}$ which evolve according to 
$\cvec{h}_t = f(\cvec{h}_{t-1}, \cvec{a}_{t-1}, \cvec{s}_{t-1})$ and $\cvec{s}_t \sim p(\cvec{s}_t | \cvec{h}_t)$. 
Here $f$ is implemented using a \emph{Gated Recurrent Unit (GRU)}~\cite{cho2014gru}. 
This results in a nonlinear, autoregressive process that cannot be parallelized over time.
Further, \emph{RSSMs} assume a filtering inference model $q(\cvec{s}_t | \cvec{h}_t, \cvec{o}_t)$, where $\cvec{h}_t$ accumulates all information from the past. 
The \emph{RSSM}'s inference scheme struggles with correctly estimating uncertainties as the resulting lower bound is not tight~\cite{becker2022uncertainty}. 
In tasks where such uncertainties are relevant, this lack of principled uncertainty estimation causes poor performance for downstream applications.

As a remedy, the \emph{Variational Recurrent Kalman Network (VRKN)}~\cite{becker2022uncertainty} builds on a linear Gaussian SSM in a latent space which allows inferring smoothed belief states $\cvec q(\cvec{z}_t | \cvec{o}_{\leq T}, \cvec{a}_{\leq T})$ required for a tight bound. 
The \emph{VRKN} removes the need for a deterministic path and improves performance under uncertainty. 
However, it linearizes the dynamics model around the mean of the filtered belief, resulting in a nonlinear autoregressive process that cannot be parallelized.

In contrast, \emph{Recall to Image (R2I)}~\cite{samsami2024mastering} builds on the \emph{RSSM} and improves computational efficiency at the cost of a more simplistic inference scheme.
It uses \emph{S4}~\cite{gu2021s4} instead of a \emph{GRU} to parameterize the deterministic path $f$ but has to remove the inference's dependency on $\cvec{h}_t$ to allow efficient parallel computation. 
The resulting inference model, $q(\cvec{z}_t | \cvec{o}_t)$ is non-recurrent and neglects all information from other time steps.
Thus, while \emph{R2I} excels on memory tasks, it is highly susceptible to noise and partial-observability as the inference cannot account for inconsistent or missing information in $\cvec{o}_t$. 

Our approach, \emph{KalMamba}, combines the tight variational lower bound of the \emph{VRKN} with a parallelizable \emph{Mamba}~\cite{gu2023mamba} backbone to learn the parameters of the dynamics.
It thus omits the nonlinear autoregressive linearization process.
Combined with our custom PyTorch routines for time-parallel filtering and smoothing~\cite{sarkka2020temporal}, this approach allows efficient training with the \emph{VRKN}s principled, uncertainty-capturing objective.

\section{KalMamba}

On a high level, \emph{KalMamba} embeds a linear Gaussian State Space Model into a latent space and learns its dynamics model's parameters using a backbone consisting of several mamba layers. 
It employs a time-parallel Kalman smoother in this latent space to infer latent beliefs for training and acting.
By exploiting the associativity of the underlying operations, we can utilize parallel scans for this parallelization. 
\emph{KalMamba} employs a tight variational lower bound objective that allows appropriate modeling of uncertainties in noisy, partial-observable systems.
We then use a \emph{Soft Actor Critic}~\cite{haarnoja2018sac} approach to learn to act, avoiding autoregressive rollouts for policy learning. 

\subsection{The \emph{KalMamba} Model}

\definecolor{steelblue31119180}{RGB}{31,119,180}
\definecolor{snekgreen}{RGB}{163,230,114}
\begin{figure*}[t]
    \centering
    \includegraphics[width=\textwidth]{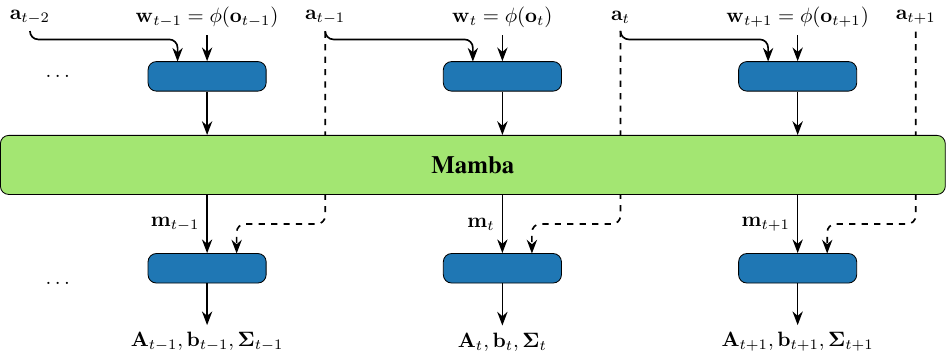}                      
    \caption{
        Schematic of the Mamba~\cite{gu2023mamba} based backbone to learn the system dynamics. 
        It shares the inference model's encoder $\phi(\cvec{o}_t)$ and intermediate representation $\cvec{w}_t$. 
        Each $\cvec{w}_t$ is then concatenated to the previous action $\cvec{a}_{t-1}$, fed through a  {\color{steelblue31119180} small Neural Network (NN)} and given to {\color{snekgreen!60!black} \emph{Mamba} model} which accumulates information over time and emits a representation $\cvec{m}_t(\cvec{o}_{t \leq}, \cvec{a}_{\leq t-1})$ containing the same information as the filtered belief $q(\cvec{z}_t | \cvec{o}_{t \leq}, \cvec{a}_{\leq t-1})$. 
        We then concatenate each $\cvec{m}_t$ with the current action $\cvec{a}_t$ and use another {\color{steelblue31119180} small NN} to compute the dynamics parameters $\cmat{A}_t, \cvec{b}_t$ and $\cmat{\Sigma}_t$.         
        This scheme allows us to use the intermediate representation $\cvec{m}_t$ for regularization and we regularize it towards the filtered belief's mean using a Mahalanobis regularizer (c.f. ~\autoref{eq:mahal_reg}).
        Finally, the {\color{steelblue31119180} small NNs} include Monte-Carlo Dropout~\cite{gal2016dropout} to model epistemic uncertainty. 
    }
    \label{fig:fig2}
\end{figure*}

To connect the original, high-dimensional observations $\cvec{o}_t$ to the latent space for inference, we introduce an intermediate auxiliary observation $\cvec{w}_t$, which is connected to the latent state by an observation model $q(\cvec{w}_t | \cvec{z}_t) = \mathcal{N}\left(\cvec{w}_t | \cvec{z}_t, \cmat{\Sigma}_t^{\cvec{w}}\right)$~\cite{becker2019recurrent, shaj2020action}. 
Here, we assume $\cvec{w}_t$ to be observable and extract it, together with the diagonal observation covariance  $\cmat{\Sigma}_t^{\cvec{w}}$ from the observation using an encoder network; $\left(\cvec{w}_t,  \cvec{\Sigma}_t^{\cvec{w}} \right) = \phi(\cvec{o}_t)$. 
This approach allows us to model the complex dependency between $\cvec{z}_t$ and  $\cvec{o}_t$ using the encoder while having a simple observation model for inference in the latent space. 
Opposed to modeling $\cvec{w}_t$ as a random variable~\cite{fraccaro2017kvae,klushyn2021latent}, modeling it is observable results in fewer latent variables which simplifies inference and allows direct propagation of the observation uncertainties from the encoder to the state.  

We parameterize the dynamics model as   
\begin{align}
p(\cvec{z}_{t+1} | \cvec{z}_{t}, \cvec{a}_{t}) = \mathcal{N}\left(\cvec{z}_{t+1} | \cmat{A}_t(\cvec{o}_{\leq t}, \cvec{a}_{\leq t})\cvec{z}_t + \cvec{b}_t(\cvec{o}_{\leq t}, \cvec{a}_{\leq t}), \cmat{\Sigma}_t^{\textrm{dyn}}(\cvec{o}_{\leq t}, \cvec{a}_{\leq t}) \right)
\label{eq:dynamics_model}
\end{align}
where both $\cmat{A}_t$ and $\cmat{\Sigma}^{\textrm{dyn}}_t$ are diagonal matrices and we constrain the (eigen)values of $\cmat{A}$ to be between $0.4$ and $0.99$.
This constraint ensures the resulting dynamics are plausible and stable.  
This approach effectively linearizes the dynamics parameters $\cmat{A}_t, \cvec{b}_t$ and $\cmat{\Sigma}^{\textrm{dyn}}_t$ around all past observations and actions.
Crucially, the resulting dynamics are linear in $\cvec{z}_t$ enabling the closed-form inference of beliefs using standard Kalman filtering and smoothing. 

Parameterizing the dynamics model of~\autoref{eq:dynamics_model} naively can lead to poor representations, as information can bypass the actual SSM through the linearization backbone.
To counter this, we design the backbone architecture as depicted in~\autoref{fig:fig2}. 
For each timestep, we concatenate $\cvec{w}_{t}$ and $\cvec{a}_{t-1}$, transform each resulting vector using a small neural network, feed it through a \emph{Mamba}~\cite{gu2023mamba} model and linearly project the output to a vector $\cvec{m}_{t}$ of the same dimension as the latent state $\cvec{z}_t$. 
Each $\cvec{m}_t$ now accumulates the same observations and actions used to form the corresponding filtered belief $q(\cvec{z}_{t} | \cvec{o}_{\leq t}, \cvec{a}_{\leq t-1})$.  
We then take $\cvec{m}_t$ and the action $\cvec{a}_t$ to compute the dynamics parameters using another small neural network. 
This bottleneck introduced by $\cvec{m}_t$ allows us to regularize the model as discussed below. 
Following~\cite{becker2022uncertainty} we further include Monte-Carlo Dropout~\cite{gal2016dropout} into this architecture, as explicitly modeling the epistemic uncertainty is crucial when working with a smoothing inference. 

The generative observation model is given by a decoder network $p(\cvec{o}_t | \cvec{z}_t)$. 
The observations are modeled as Gaussian with learned mean and fixed standard deviation. 
Finally, we assume an initial state distribution $p(\cvec{z}_0)$ that is a zero mean Gaussian with a learned variance $\cmat{\Sigma}_0$. 

Given the latent observation model $q(\cvec{w}_t | \cvec{z}_t)$, and the shared, pre-computable, linear dynamics model, we can efficiently infer belief states using extended Kalman filtering and smoothing. 
Recent work~\citep{sarkka2020temporal} shows how to formulate such filtering and smoothing as associative operations amenable to temporal parallelization using associative scans. 
We implement these operations in PyTorch~\cite{paszke2019pytorch}.
Similar to \emph{S5}~\citep{smith2022simplified} or \emph{Mamba}~\citep{gu2023mamba} this implementation yields a logarithmic time complexity, given sufficiently many parallel cores. 
Additionally, as the dynamics matrix $\cmat{A}_t$ and all model covariances, i.e., $\cmat{\Sigma}_t^{\textrm{dym}}$, $\cmat{\Sigma}_t^{\textrm{obs}}$, and $\cmat{\Sigma}_0$, are diagonal, the same holds for the covariances of the filtered and smoothed beliefs.
Thus we can replace costly matrix operations during Kalman filtering and smoothing with point-wise operations, which further ensures \emph{KalMamba's} efficiency. 

\subsection{Training the Model}
After inserting the state space assumptions of our generative and inference models, the standard variational lower bound to the data marginal log-likelihood~\cite{kingma2013auto} for a single sequence simplifies to~\cite{becker2022uncertainty}
\begin{align*}
	\mathcal{L}_{\textrm{ssm}}(\cvec{o}_{\leq T}, \cvec{a}_{\leq T}) =\sum_{t=1}^T \biggl(  & \mathbb{E}_{q(\cvec{z}_t | \cvec{o}_{\leq T}, \cvec{a}_{\leq T})} \left[ \log p(\cvec{o}_t | \cvec{z}_t) \right ]  - \\ & \mathbb{E}_{q(\cvec{z}_{t-1} | \cvec{o}_{\leq T}, \cvec{a}_{\leq T})} \left[ \KL{q(\cvec{z}_{t} | \cvec{z}_{t-1}, \cvec{a}_{\geq t-1}, \cvec{o}_{\geq t})}{p(\cvec{z}_{t} | \cvec{z}_{t-1}, \cvec{a}_{t-1})}\right] \biggr).  \nonumber% \label{eq::ssm_elbo}
\end{align*}
Due to the smoothing inference, this lower bound is tight and allows accurate modeling of the underlying system's uncertainties. 
To evaluate the lower bound we need the smoothed dynamics $q(\cvec{z}_{t} | \cvec{z}_{t-1}, \cvec{a}_{\geq t-1}, \cvec{o}_{\geq t})$ whose parameters we can compute given the equations provided in \cite{becker2022uncertainty}. 

To regularize the \emph{Mamba}-based backbone used to learn the dynamics, we  incentivize $\cvec{m}_t$ to correspond to the filtered mean using a Mahalonobis distance
\begin{align}
R(\cvec{o}_{\leq T}, \cvec{a}_{\leq{T}}) = \sum_{t=1}^T \left( \cvec{m}_t(\cvec{o}_{\leq{t}}, \cvec{a}_{\leq{t -1}} - \cvec{\mu}_t^+ \right) ^T \left(\cmat{\Sigma}_t^{+}\right)^{ -1}\left(\cvec{m}_t(\cvec{o}_{\leq{t}}, \cvec{a}_{\leq{t -1}}) - \cvec{\mu}_t^+\right) ,
\label{eq:mahal_reg}
\end{align}
$\cvec{\mu}_t^+$ and $\cmat{\Sigma}_t^+$ denote the mean and variance of the filtered belief $q(\cvec{z}_{t} | \cvec{o}_{\leq{t}}, \cvec{a}_{\leq{t -1}})$.
This regularization discourages the model from bypassing information over the \emph{Mamba} backbone.
This mirrors many established models such as the classical extend Kalman Filter~\cite{jazwinski1970stochastic}, which linearize directly around this mean, but still allows associative parallel scanning. 

Finally, we add a reward model $p(r_t | \cvec{z}_t)$, predicting the current reward from the latent state using a small neural network.
While this is not strictly necessary for standard policy learning on top of the representation, it nevertheless helps the model to focus on task-relevant details and learn a good representation for control~\cite{srivastava2021core,tomar2023whatmatter}.
Including this reward term and the Mahalonobis regularize, the full maximization objective for a single sequence is given as
\begin{align*}
  \mathcal{L}_{\textrm{KalMamba}}(\cvec{o}_{\leq T}, \cvec{a}_{\leq T}) =  \mathcal{L}_{\textrm{ssm}}(\cvec{o}_{\leq T}, \cvec{a}_{\leq T}) +  \mathbb{E}_{q(\cvec{z}_t | \cvec{o}_{\leq T}, \cvec{a}_{\leq T})} \left[ \log p(r_t | \cvec{z}_t) \right ] - \alpha R(\cvec{o}_{\leq T}, \cvec{a}_{\leq T}). 
\end{align*}

\subsection{Using \emph{KalMamba} for Reinforcement Learning} 
We learn a policy on top of the \emph{KalMamba} state space representation using \emph{Soft Actor Critic (SAC)}~\cite{haarnoja2018sac}. 
Here, we use the mean of the variational filtered belief $q(\cvec{z}_t | \cvec{o}_{\leq t}, \cvec{a}_{\leq t-1})$ as input for the actor and, together with the action $\cvec{a}_t$ for the critic. 
Importantly, we cannot smooth during acting as future observations and actions are unavailable. 
However, while not directly involved in the loss, the filter belief is still meaningful as the smoothing pass introduces no additional parameters.
This inductive bias induces a tight coupling between filtered and smoothed belief that ensures the reasonableness of the former. 
We independently train the \emph{KalMamba} world model and \textit{SAC} by stopping the actor's and critic's gradients from propagating through the world model. 
We use \emph{SAC} instead of the typical latent imagination strategy used with \emph{RSSMs}, the \emph{VRKN} and \emph{R2I}.
For all $4$ models, rolling out policies in the latent space is autoregressive but these rollouts can be avoided by using a $Q$-function directly on the inferred belief states. 

%\newpage
\section{Experiments}
\begin{figure*}[t]
\centering
    \begin{minipage}{\textwidth}
    \centering
    \includegraphics{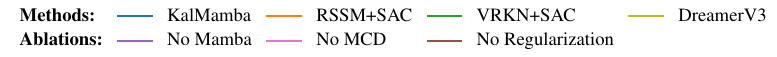}
    %\tikzsetnextfilename{01_legend_combined}
    %\input{figures/quantitative/tikz/images/legend_combined}
    \end{minipage}
    \begin{minipage}{\textwidth}
    \includegraphics{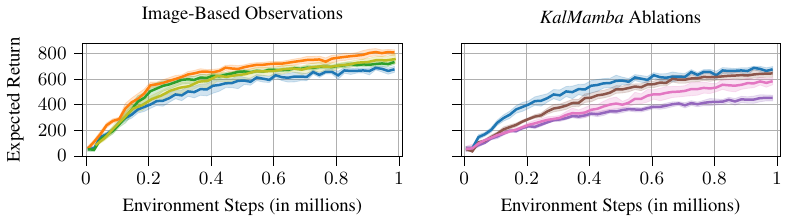}
    %\tikzsetnextfilename{02_aggregate_combined}
    %\input{figures/quantitative/tikz/images/aggregate_combined}
    \end{minipage}%
    \caption{
    Aggregated expected returns for image-based observations.
    (\textbf{Left})~\ours~is slightly worse but overall competitive with the different baselines.
    Combining either baseline SSM with SAC matches or exceeds the performance of \textit{DreamerV3}.
%\todo{intuition?}
    (\textbf{Right})
    Using \textit{Mamba} to learn the dynamics is crucial for good model performance. 
    Similarly, both Monte-Carlo Dropout and the regularization loss of Equation~\ref{eq:mahal_reg} stabilize the training process and lead to higher expected returns.
    }
    \label{fig:quantitative_images_ablations}
    % \vspace{-0.2cm}
\end{figure*}
We evaluate \emph{KalMamba} on $4$ tasks from the DeepMind Control (DMC) Suite, namely \texttt{cartpule\_swingup, quardruped\_walk, walker\_walk,} and \texttt{walker\_run}. 
We train each task for $1$ million environment steps with sequences of length $32$ and run $20$ evaluation runs every $20,000$ steps.
We report the expected return using the mean and $95\%$ stratified bootstrapped confidence intervals~\cite{agarwal2021deep} for $4$ seeds per environment.
\autoref{app_sec:hps} provides all hyperparameters.
\autoref{app_sec:additional_results} provides per-task results for all experiments.

We compare against \emph{Recurrent State Space Models (RSSM)} and the \emph{Variational Recurrent Kalman Network (VRKN)}.
To isolate the effect of the SSMs' representations, we combine both with \emph{SAC}~\cite{haarnoja2018sac} as the RL algorithm, instead of using latent imagination~\cite{hafner2019dream}. 

\subsection{Standard Image Based Tasks}
%\begin{wrapfigure}{R}{0.5\textwidth}
\begin{figure}[t]
\centering
\begin{minipage}{0.5\textwidth}
\includegraphics{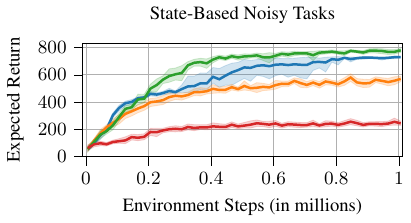}
\end{minipage}%
\begin{minipage}{0.2\textwidth}
\vspace{-1.5cm}
\includegraphics{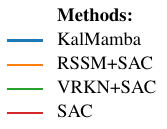}
\end{minipage}%
\caption{
Aggregated expected returns for the state-based noisy tasks. 
\emph{KalMamba} clearly outperforms the \emph{RSSM} while almost matching the \emph{VRKN}'s performance.
Naively using \emph{SAC} is insufficient, which testifies to the increased difficulty due to the noise. }
\label{fig:quantitativestates}
\end{figure}

We first compare on standard image-based observations of the different tasks and include \emph{DreamerV3}~\cite{hafner2023mastering} results for reference. 
The left side of 
\autoref{fig:quantitative_images_ablations} shows the aggregated expected returns.
The results indicate that~\ours~is slightly worse, but overall competitive to the two baseline SSMs and \textit{DreamerV3}, while being parallelizable and thus much more efficient to train.
Interestingly, both SSMs work well when combined with \emph{SAC}, matching or outperforming \textit{DreamerV3}. 
%\todo{intuition?}

%\todo{move ablations further down?}
We also conduct ablations for some of the main design choices of \emph{KalMamba} on the right side of~\autoref{fig:quantitative_images_ablations}.
\emph{No Mamba} removes the Mamba layers from the dynamics backbone in~\autoref{fig:fig2}.
Similar to the selection mechanism of \emph{Mamba}~\cite{gu2023mamba} itself, the resulting approach linearizes the dynamics around the current action and observation, instead of all previous observations and actions. 
The results show that this is insufficient for \emph{KalMamba}, presumably because it uses only a single SSM layer instead of the stacked layers used by \emph{Mamba}. 
Furthermore, \emph{No Regularization loss} removes the Mahalanobis regularization from the model and \emph{No Monte Carlo Dropout} removes Monte-Carlo Dropout from the dynamics backbone.
Here, the results indicate that regularizing $\mathbf{m}_t$ and explicitly modeling the epistemic uncertainty are crucial for \emph{KalMamba's} performance. 

\subsection{Low Dimensional Tasks with Observation and Dynamics Noise}
To test the models' capabilities under uncertainties, we use the state-based versions of the tasks and add both observation and dynamics noise. 
The observation noise is sampled from $\mathcal{N}(0, 0.3)$ and added to the observation.
The dynamics noise is also sampled from $\mathcal{N}(0, 0.3)$ and added to the action before execution.
However, unlike exploration noise, this addition happens inside the environment and is invisible to the world model and the policy.
We include \emph{SAC} without a world model in our experiments as a baseline to evaluate the difficulty of the resulting tasks. % to show that the resulting tasks are too complex to be solved naively. 

The results in \autoref{fig:quantitativestates} show that naively using \emph{SAC} fails in the presence of noise. 
While the \emph{RSSM} manages to improve performance it is still significantly outperformed by \emph{VRKN} and \emph{KalMamba}, which both use the robust smoothing inference scheme.
\emph{KalMamba} needs slightly more environment steps to converge but ultimately almost matches the \emph{VRKN}'s performance while being significantly faster to run. 

\subsection{Runtime Analysis}
\begin{figure*}[t]
\centering
    \begin{minipage}{\textwidth}
    \centering
    \includegraphics{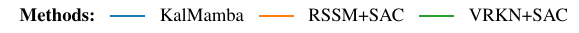}
    \end{minipage}
    \begin{minipage}{\textwidth}
    \centering
    \includegraphics{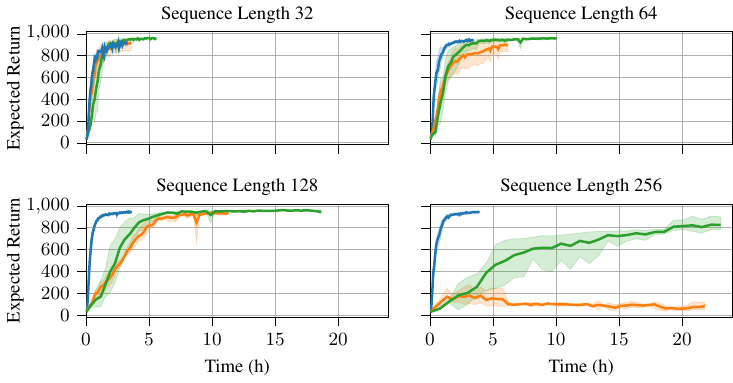}
    \end{minipage}%
    \caption{
    Wall-clock time evaluations on the state-based noisy \texttt{walker-walk} for~\ours~, the \emph{RSSM}, and the \emph{VRKN} for different training context lengths for $1$ million environment steps or up to $24$ hours.
    This time limitation only affected the \emph{VRKN} training for $256$ steps, which reached $650$ thousand steps after $24$ hours. 
    While all methods work well for short sequences of length $32$ (\textbf{Top Left}), the efficient parallelization of~\ours~allows it to scale gracefully to and even improve performance for longer sequences of up to $256$ steps, where the other methods fail (\textbf{Bottom Right}).
    }
    \label{fig:quantitative_timings_length}
    % \vspace{-0.2cm}
\end{figure*}
%\todo{
%Compared to existing probabilistic SSMs, \ours~allows for efficient parallel inference thanks to associative scans~\citep{sarkka2020temporal}.
%}
To show the benefit of \emph{KalMambas} efficient parallelization using associative scans, we compare its wall-clock runtime to that of the SSM baselines on the state-based noisy version of \texttt{walker-walk} for training sequences of increasing length.
The models share a PyTorch implementation and differ only in the SSM. 
We run each experiment on a single Nvidia Tesla H100 GPU, for up to $1$ million steps or $24$ hours.
Figure~\ref{fig:quantitative_timings_length} shows the resulting expected returns.
While all models work well for the short sequences of length $32$ that are used for the main results above, the training time of the baseline SSMs scales linearly with the sequence length, causing slower convergence and a time-out after $24$ hours and $650$ thousand environment steps for the \emph{VRKN} for a length of $256$.
In comparison, \ours~shows negligible additional training cost for increased sequence lengths.
Further, while the absolute performance of both baselines decreases as the training sequences get longer,~\ours~slightly improves performance when trained on more than $32$ steps.
These results indicate that~\ours~efficiently utilizes long-term context information through its \emph{Mamba} backbone, whereas the dynamics models of the baseline SSMs have difficulty with too-long training sequences.

\begin{figure}[t]
    \centering
    % \begin{minipage}{0.9\textwidth}
    \begin{minipage}{\textwidth}
    \centering
    \includegraphics{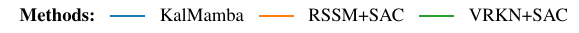}
    \end{minipage}
    \begin{minipage}{\textwidth}
    \centering
    \includegraphics{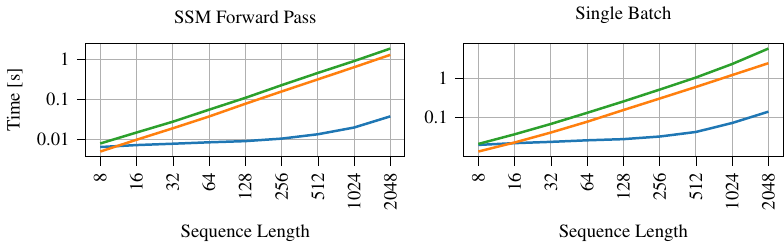}
    \end{minipage}

    % \end{minipage}%
    \hspace{0.03\textwidth}%
    \caption{
    Runtime comparison of~\ours, the RSSM and the VRKN for (\textbf{Left}) a SSM forward pass and (\textbf{Right}) a single training batch.
    While the computational cost of both baseline models scales linearly in the sequence length, \ours~utilizes associative scans for efficient parallelism and thus near-logarithmic runtime on modern accelerator hardware.
    }
    \label{fig:runtime}
\end{figure}

Investigating this further, we visualize the wall-clock time of a single SSM forward pass and a single training batch for different sequence lengths in Figure~\ref{fig:runtime}.
While both the \textit{RSSM} and \textit{VRKN} scale linearly with the sequence length,~\ours~shows near-logarithmic scaling even for longer sequences thanks to its efficient parallelism. 
We expect further significant speedups for \emph{KalMamba} with a potential custom CUDA implementation, similar to \emph{Mamba}.

%\newpage
\section{Conclusion}
We proposed \emph{KalMamba}, an efficient State Space Model (SSM) for Reinforcement Learning (RL) under uncertainty. 
It combines the uncertainty awareness of probabilistic SSMs with recent deterministic SSMs' scalability by embedding a linear Gaussian SSM into a latent space.
We use \emph{Mamba}~\cite{gu2023mamba} to learn the linearized dynamics in this latent space efficiently.
Inference in this SSM amounts to standard Kalman filtering and smoothing and is amenable to full parallelization using associative scans~\cite{sarkka2020temporal}.
During model learning, this allows time-parallel estimation of smoothed belief states, which allows the efficient usage of principled objectives for uncertainty estimation, especially over long sequences. 

Our experiments on low-dimensional states and image observations indicate that~\emph{KalMamba} can match the performance of state-of-the-art stochastic SSMs for RL under uncertainty.
In terms of both runtime and performance, \ours~scales more gracefully to longer training sequences.
In particular, its performance improves with sequence length while it degrades for the baseline SSMs.

\textbf{Limitations and Future Work.}
The present work explores \emph{KalMamba}'s potential in small-scale experiments, but a more elaborate evaluation of diverse, more realistic tasks would help to explore our method's strengths and weaknesses.
A thorough comparison of recent baselines is needed to contextualize \ours~against existing time-efficient SSMs with simplified, non-smoothing inference schemes~\cite{samsami2024mastering}.
We additionally aim to refine \emph{KalMamba} to improve its performance across wide-ranging tasks.
In this context, we plan to model the state with a complex-valued random variable to expand the range of dynamics models that can be learned.
Other ideas include improving the regularization of the Mamba backbone and investigating more advanced policy learning methods that make use of the uncertainty in the filtered beliefs.

\bibliography{main}
\bibliographystyle{abbrv}

%%%%%%%%%%%%%%%%%%%%%%%%%%%%%%%%%%%%%%%%%%%%%%%%%%%%%%%%%%%%
\clearpage
\appendix
\section{Hyperparameters and Implementation Details}
\label{app_sec:hps}
\autoref{tab:wm_hp} lists all hyperparameters of the \emph{KalMamba} model and \autoref{tab:sac_hp} lists the hyperparameters of \emph{Soft Actor Critic (SAC)}~\cite{haarnoja2018sac} used for control. 
\begin{table}[ht!]
    \caption{World Model Hyperparameters}
    \label{tab:wm_hp}
    \centering
    \begin{tabular}{lcc}
    \toprule    
    Hyperparameter & Low Dimensional DMC & Image Based DMC \\
    \midrule
    \multicolumn{3}{c}{World Model}\\
    \midrule
    Encoder & $2 \times 256$ Unit NN with ELU  & ConvNet from \cite{ha2018world, hafner2019dream} with ReLU \\
    Decoder & $2 \times 256$ Unit NN with ELU  & ConvNet from \cite{ha2018world, hafner2019dream} with ReLU \\    
    Reward Decoder & \multicolumn{2}{c}{$2 \times 256$ Unit NN with ELU} \\ 
    Latent Space Size &  \multicolumn{2}{c}{$230$ ($30$ Stoch. + $200$ Det. for RSSM}\\
    \midrule
    \multicolumn{3}{c}{Mamba Backbone}\\
    \midrule
    num blocks & \multicolumn{2}{c}{$2$} \\ 
    d\_model & \multicolumn{2}{c}{256} \\ 
    d\_state & \multicolumn{2}{c}{64} \\ 
    d\_conv & \multicolumn{2}{c}{2} \\ 
    dropout probability & \multicolumn{2}{c}{0.1} \\ 
    activation & \multicolumn{2}{c}{SiLU} \\ 
    pre mamba layers &   \multicolumn{2}{c}{ $2 \times 256$ Unit NN with SiLU } \\ 
    post mamba layers &  \multicolumn{2}{c}{VRKN Dynamics Model Architecture from \cite{becker2022uncertainty} with SiLU} \\ 
    \midrule
    \multicolumn{3}{c}{Loss}\\
    \midrule
    KL Balancing &  \multicolumn{2}{c}{$0.8$ for RSSM, $0.5$ for VRKN, KalMamba} \\
    Free Nats &  \multicolumn{2}{c}{$3$} \\ 
    $\alpha$ (regularization scale)  & \multicolumn{2}{c}{$1$, KalMamba only} \\ 
    \midrule
    \multicolumn{3}{c}{Optimizer (Adam~\cite{kingma2015adam})}\\
    \midrule
    Learning Rate & \multicolumn{2}{c}{$3 \cdot 10^{-4}$} \\    

    \bottomrule
    \end{tabular}
\end{table}

\begin{table}[ht!]
    \caption{SAC Hyperparameters}
    \label{tab:sac_hp}
    \centering
    \begin{tabular}{lcc}
    \toprule    
    Hyperparameter & Low Dimensional DMC & Image Based DMC \\
    \midrule
    Actor-Network & $2 \times 256$ Unit NN with ReLU  & $3 \times 1024$ Unit NN with ELU  \\
    Critic-Network & $2 \times 256$ Unit NN with ReLU  & $3 \times 1024$ Unit NN with ELU  \\
    \midrule
    Actor Optimizer  & \multicolumn{2}{c}{Adam with learning rate $3 \times 10^{-4}$} \\ 
    Critic Optimizer  & \multicolumn{2}{c}{Adam with learning rate $3 \times 10^{-4}$} \\ 
    \midrule
    Target Critic Update Fraction  & \multicolumn{2}{c}{0.005} \\ 
    Target Critic Update Interval  & \multicolumn{2}{c}{1} \\ 
    \midrule
    Target Entropy     & \multicolumn{2}{c}{$-d_{\textrm{action}}$} \\ 
    Entropy Optimizer & \multicolumn{2}{c}{Adam with learning rate $3 \times 10^{-4}$} \\ 
    Initial Learning Rate & \multicolumn{2}{c}{$0.1$} \\ 
    \midrule
    discount $\gamma$ & \multicolumn{2}{c}{$0.99$} \\ 
    \bottomrule
    \end{tabular}
\end{table}

\subsection{Baselines.}
Both \emph{RSSM+SAC} and \emph{VRKN+SAC} use the same hyperparameters as \emph{KalMamba} where applicable.
For all other hyperparameters, we use the defaults from \cite{hafner2019dream} and \cite{becker2022uncertainty} respectively. 
The \emph{SAC} baseline uses the hyperparameters listed in \autoref{tab:sac_hp} and the results for \emph{DreamerV3}~\cite{hafner2023mastering} are provided by the authors\footnote{\url{https://github.com/danijar/dreamerv3}}. 
\section{Additional Results}
\label{app_sec:additional_results}
\begin{figure*}[h!]
\centering
    \begin{minipage}{\textwidth}
    \centering
    \includegraphics{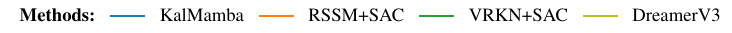}
    \end{minipage}    
    \begin{minipage}{\textwidth}
    \includegraphics{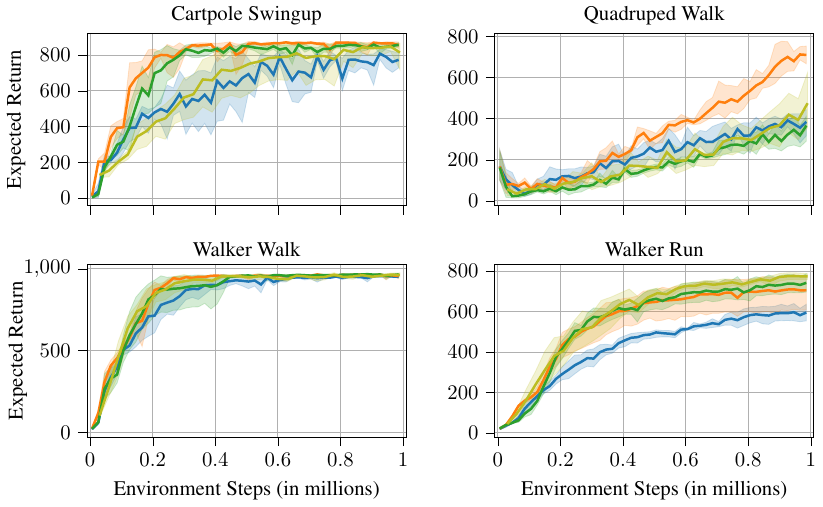}
    \end{minipage}%
    \caption{
    Task-wise evaluations of the DeepMind Control Suite on image-based observations.
    Dreamer-v$3$ shows a performance similar to RSSM+SAC.
    %\todo{Caption.}
    }
    \label{app_fig:quantitative_images_grid}
    % \vspace{-0.2cm}
\end{figure*}

\begin{figure*}[t]
\centering
    \begin{minipage}{\textwidth}
    \centering
    \includegraphics{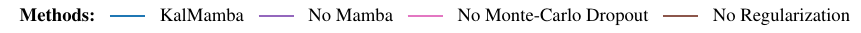}
    \end{minipage}    
    \begin{minipage}{\textwidth}
    \includegraphics{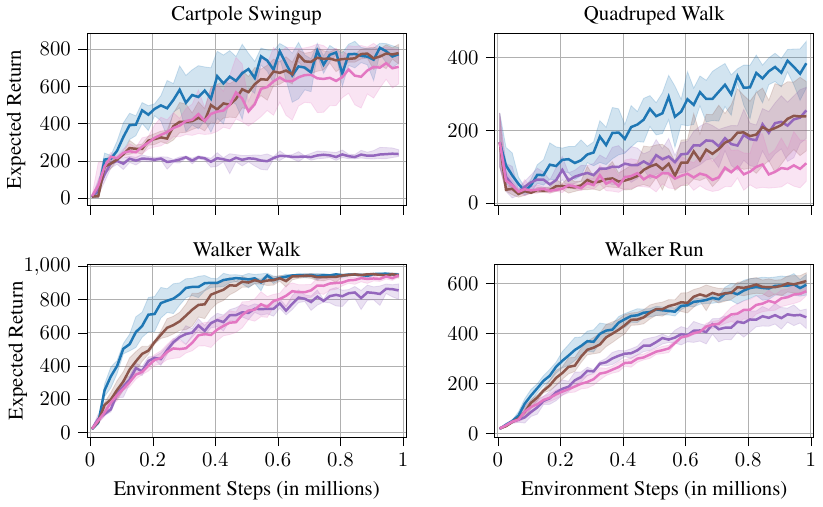}
    \end{minipage}%
    \caption{
    Task-wise evaluations of the DeepMind Control Suite for different~\ours~ablations.
    Monte-Carlo Dropout and the Mahalanobis regularization make the largest difference for the hardest task in the suite, i.e., \texttt{quadruped\_walk}.
    }
    \label{app_fig:quantitative_images_ablations}
    % \vspace{-0.2cm}
\end{figure*}

\begin{figure*}[t]
\centering
    \begin{minipage}{\textwidth}
    \centering
    \includegraphics{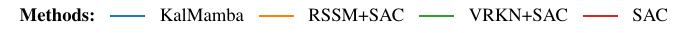}
    \end{minipage}   
    \begin{minipage}{\textwidth}
    \includegraphics{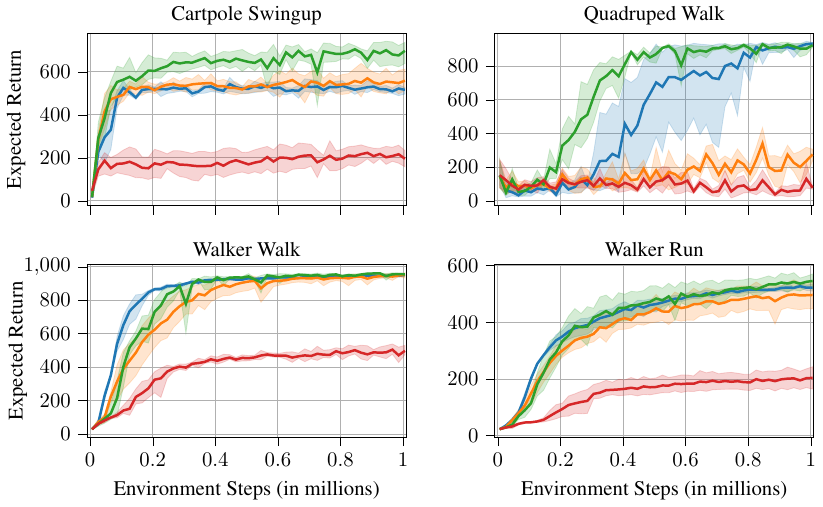}
    \end{minipage}%
    \caption{
    Task-wise evaluations of the DeepMind Control Suite on low-dimensional state representations.
    \ours~performs on par with or better than the RSSM on all tasks, and is only outperformed by the computationally more expensive VRKN on \texttt{cartpole\_ swingup}.
    }
    \label{app_fig:quantitative_states_grid}
\end{figure*}

We provide results for the individual tasks of the Deepmind Control Suite for 
image-based observations in ~\autoref{app_fig:quantitative_images_grid} and the different~\ours~ablations in ~\autoref{app_fig:quantitative_images_ablations}.
\autoref{app_fig:quantitative_states_grid}, shows the per-task results for the noisy state-based environments.
%{\color{red} Add Derivations from VRKN paper?}

%%%%%%%%%%%%%%%%%%%%%%%%%%%%%%%%%%%%%%%%%%%%%%%%%%%%%%%%%%%%

\end{document}